\documentclass{llncs}
\usepackage{amssymb}

\usepackage{url}
\usepackage{listings}
\usepackage{graphicx}
\usepackage{color}
\usepackage{epsfig}

\newcommand{\knowrex}{KnowRex~}
\newcommand{\knowrexx}{KnowRex}

\newcommand{\verba}{\vspace{-0.7mm}}
\newcommand{\mytt}[1]{{\small \texttt{#1}}}

\begin{document}

\title{Ontology-driven Information Extraction}

\author{Weronika T. Adrian\inst{1,2}
\and Nicola Leone\inst{1}
\and Marco Manna\inst{1}}

\institute{Department of Mathematics and Computer Science, University of Calabria, Italy
\and
AGH University of Science and Technology, al.A.Mickiewicza 30, Krakow, Poland}

\maketitle

\begin{abstract}
\emph{Homogeneous unstructured data} (HUD) are collections
of unstructured documents that share common properties, such as similar layout,
common file format, or common domain of values.
Building on such properties, it would be desirable to automatically process
HUD to access the main information through a
semantic layer -- typically an ontology -- called \emph{semantic view}.
Hence, we propose an ontology-based approach for extracting semantically
rich information from HUD, by integrating and extending recent technologies
and results from the fields of classical information extraction,
table recognition, ontologies, text annotation, and logic programming.
Moreover, we design and implement a system, named KnowRex,
that has been successfully applied to curriculum vitae in the Europass style
to offer a semantic view of them, and be able, for example,
to select those which exhibit required skills.
\end{abstract}

\section{Introduction}

\paragraph{\textbf{\emph{Context and Motivation.}}}
By its nature, the Web has been conceived
as an enormous distributed source of information
which behaves as an open system to facilitate data sharing.
However, the concrete way how the Web has been populated
gave rise to a large amount of knowledge
which is accessible only to humans but not to computers.
A large slice of this knowledge is destined to remain only human-readable.
But there is another relevant portion of it which could be
automatically manipulated to be processed by computers.
This is the case, for example, of \emph{homogeneous unstructured data} (HUD) which
are collections of unstructured documents that share
common properties, such as similar layout, common file format,
or common domain of values, just to mention a few.

Building on their common properties, it would be desirable to automatically
process HUD to access the main information
they contain through a semantic layer which is typically given in the form of an ontology,
and that we call \emph{semantic view}.
The problem of identifying and extracting information from unstructured
documents is widely studied in the field of information and knowledge management,
and is referred to as Information Extraction
(IE)~\cite{FeldmanAumann02,ChangKayedGirgisShaalan06,jiang2012ie,balke2012ie}.
However, most of the existing approaches to IE are mainly syntactic,
and do not offer a uniform, clear, and semantic view of the relevant information.


%

\paragraph{\textbf{\emph{Contribution.}}}
To offer a semantic view of a collection of HUD (even if encoded as pdf files),
we propose and implement a system, named KnowRex,
which splits the entire process in two different phases,
called \emph{design} and \emph{runtime}.
%
During the first one, the designer
$(i)$ defines the target schema for the semantic view of the original data,
$(ii)$ fixes an object model to offer a structured representation of the documents,
$(iii)$ arranges a suite of annotation units
   (such as named entity extractors, natural language processing tools, and
   annotation tools based on thesauri or regular expressions)
   to define the ``leaves'' of the object model,
$(iv)$ chooses and calibrates one of the software programs that
    partition unstructured documents into two-dimensional grids, and
$(v)$ provides formal rules to structure the documents, and to construct the
semantic view.
During the runtime phase,
the system processes the documents as prescribed in
the design phase to instantiate the object model first, and then the target schema.
In particular, the process that provides a structured representation of the documents
can be thought as a kind of IE task which is heavily
driven by domain knowledge and semantics,
while the process that constructs the semantic view from
the structured version of the documents
takes care of reorganizing the extracted knowledge to facilitate data analysis.
To sharpen our system, we
considered curriculum vitae in the \emph{Europass} style
to offer a semantic view of them and be able,
for example, to select those which exhibit required skills.
The main contributions of the paper are:

\noindent  \ $\blacktriangleright$ We present an ontology-based approach to IE
    which allows for extracting semantically rich information
   from unstructured data sharing some common features. To this end, we integrate and extend
   recent technologies and results from the fields of {\em classical information extraction},
   {\em table recognition}, {\em ontologies},
   {\em text annotation}, and {\em logic programming}.

\noindent \ $\blacktriangleright$ We design and implement a system, named KnowRex,
    which realizes our ontology-based approach to IE, and provides access
    to HUD via semantic views.

\noindent \ $\blacktriangleright$ On the application side,
   we have successfully applied KnowRex to offer a semantic view to
   curriculum vitae in the Europass style.

\paragraph{\textbf{\emph{Related Work.}}}
The literature of the academic and commercial worlds
offers a variety of approaches and tools to IE
that are either programmed manually, or learned semi-automatically.
%
A main shortcoming of these approaches, however, is their lack of
understanding of extracted information.
More recently, some works have
shown the promise of deducing and encoding
formal knowledge in the form of
ontologies~\cite{Karkaletsis2011,soner2012,anantha2013,furche2011}.
These approaches use ontologies either to improve the extraction phase
as a way to present the results of the extraction,
or to allow matching different representations across sources.
Our approach follows the line of combining several techniques
to obtain comprehensive results~\cite{mo2012,Chen2013}.
%
To better recognize named entities, we have utilized StanfordNER ({\small \url{http://nlp.stanford.edu/software/CRF-NER.shtml}}), although during the tests, we also evaluated Alchemy ({\small \url{http://www.alchemyapi.com}}),
 DBpedia Spotlight ({\small \url{http://dbpedia.org/spotlight}}),
 Extractiv ({\small \url{http://extractiv.com}}),
 OpenCalais ({\small \url{http://www.opencalais.com}}), and
 Lupedia ({\small \url{http://www.old.ontotext.com/lupedia}}).
The notion of semantic descriptors introduced in
Section~\ref{sec:knowrex-design} has been inherited from HiLeX~\cite{MORA12}.
However, we have refined and extended their shape.

\section{System Overview}
\label{sec:knowrex}

\knowrex is a \emph{framework} that allows to develop systems for \emph{Semantic Information Extraction} (i.e., information extraction based on the meaning of data).
In our approach, what drives the whole process is a \emph{semantic view} of the input data.
We start developing a new project with \knowrex by deciding what information we want to obtain in the end, and how we want to organize it.
%
It is often possible to semi-formally model the organization of data within a document (e.g. the DOM model for (X)HTML an XML languages). 
In \knowrexx, however, we take a step further and allow users to define the final semantic view that is independent from the initial structure.
This approach is closer to practical use cases, in which specialists are asked to populate \emph{existing} knowledge (or data) bases by extracting appropriate information from a collection of documents.

\begin{figure*}[h]
\begin{center}
\includegraphics
[scale=0.4]
{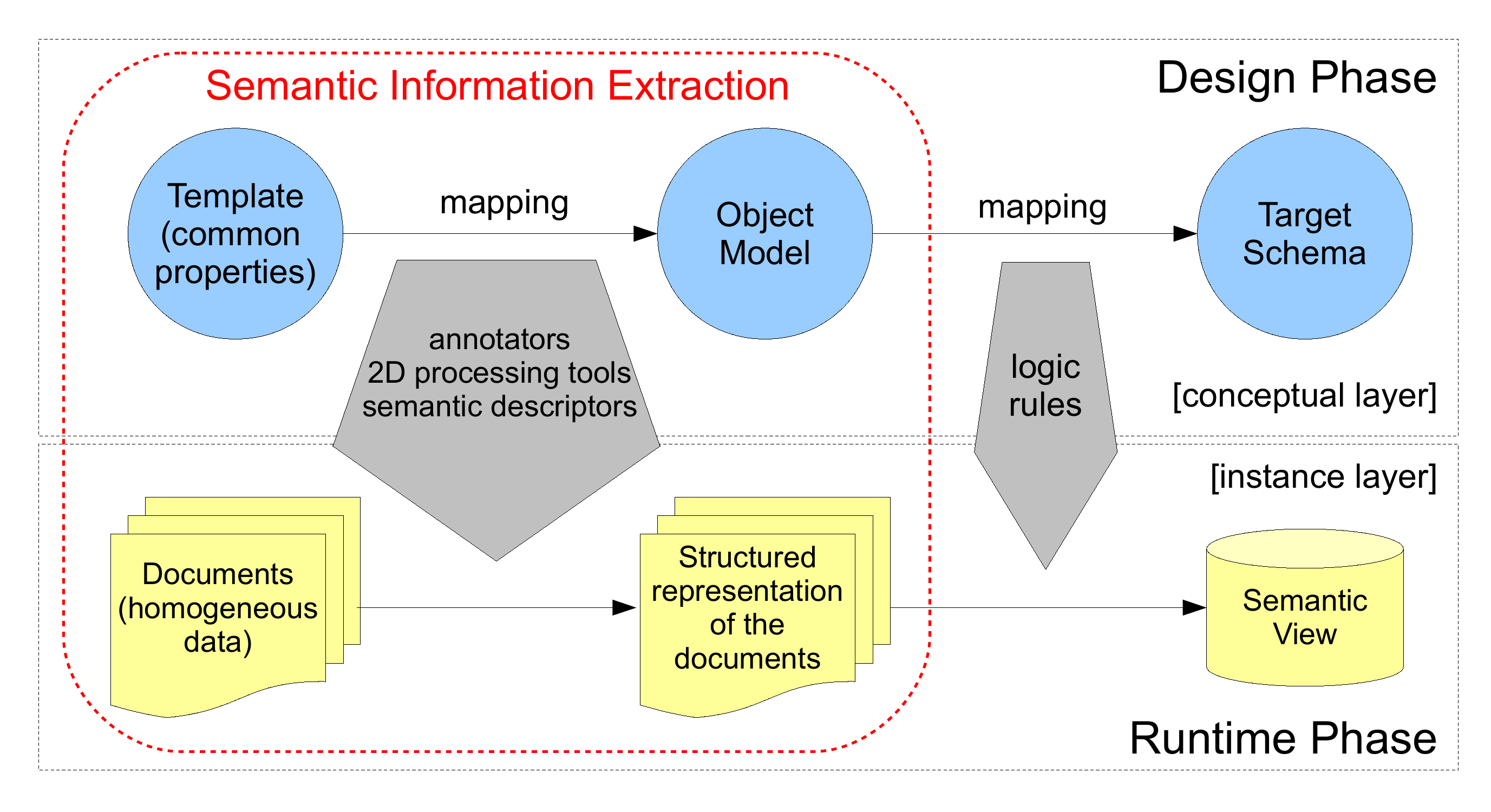}
\caption{Semantic Information Extraction with \knowrex}
\label{fig:knowrex-overview}
\end{center}
\end{figure*}

When we consider \emph{homogeneous unstructured data}, we can assume that input documents (in a given collection) share some specific features, hereafter called a ``template''.
Based on a template one can define an \emph{object model} that will capture the sort of data contained within the documents and, to some extend, the way it is organized. 
If there exists a model for a template, then the information extraction from a set of documents complying to this template will populate the object model with instances from each input document.
%
The other important part of the process is formulating a mapping from the object model to a target schema.
Such a mapping allows to reorganize objects extracted from the documents and transform them into instances of the desired semantic view (see Figure~\ref{fig:knowrex-overview}).

It is possible to define more semantic views for the same collection of data (e.g. for different use cases).
One can even imagine defining several object models for the same input and target schema.
\knowrex framework ensures flexibility in these respects, by separating the stages of extraction and the mapping to target schema and enabling reuse of the components.

Within \knowrexx, several tools and techniques have been used,
namely:
\begin{enumerate}
\item a \emph{bi-dimensional processor} for recognizing structural elements of documents,
\item one- and bi-dimensional \emph{tokenizers} for identifying basic elements of text, 
\item \emph{annotators} (third-party semantic annotators, natural language processing tools, pattern recognition tools etc.) that label single words or phrases as belonging to particular categories,
\item \emph{semantic descriptors} that allow to build the object model from the objects obtained from structural and semantic analysis, and 
\item \emph{logical rules} that allow to formulate a mapping between knowledge representations (the object model and the target schema). 
\end{enumerate}

Deployment of a new project with \knowrex is relatively easy and consists in adapting the core of the system and the external tools to work on specific data to obtain desired results.
Development of a new project is divided into  two phases: \emph{design} and \emph{runtime}.
In the former, the designer works on a conceptual level 
defining the object model (by assuming a certain template) and the target schema, and by setting the tools and writing rules that will govern the data and information transformations.
For the extraction step, the bi-dimensional processing tools, the annotators and the descriptors must be adapted, and for the mapping to target schema, logic rules must be defined.
These design choices, described in details in Section~\ref{sec:knowrex-design}, are materialized and applied in the runtime phase to extract information from the actual documents and populate semantic view(s) defined for them.
%
%

\begin{figure*}[h]
\begin{center}
\includegraphics
[scale=0.4]
{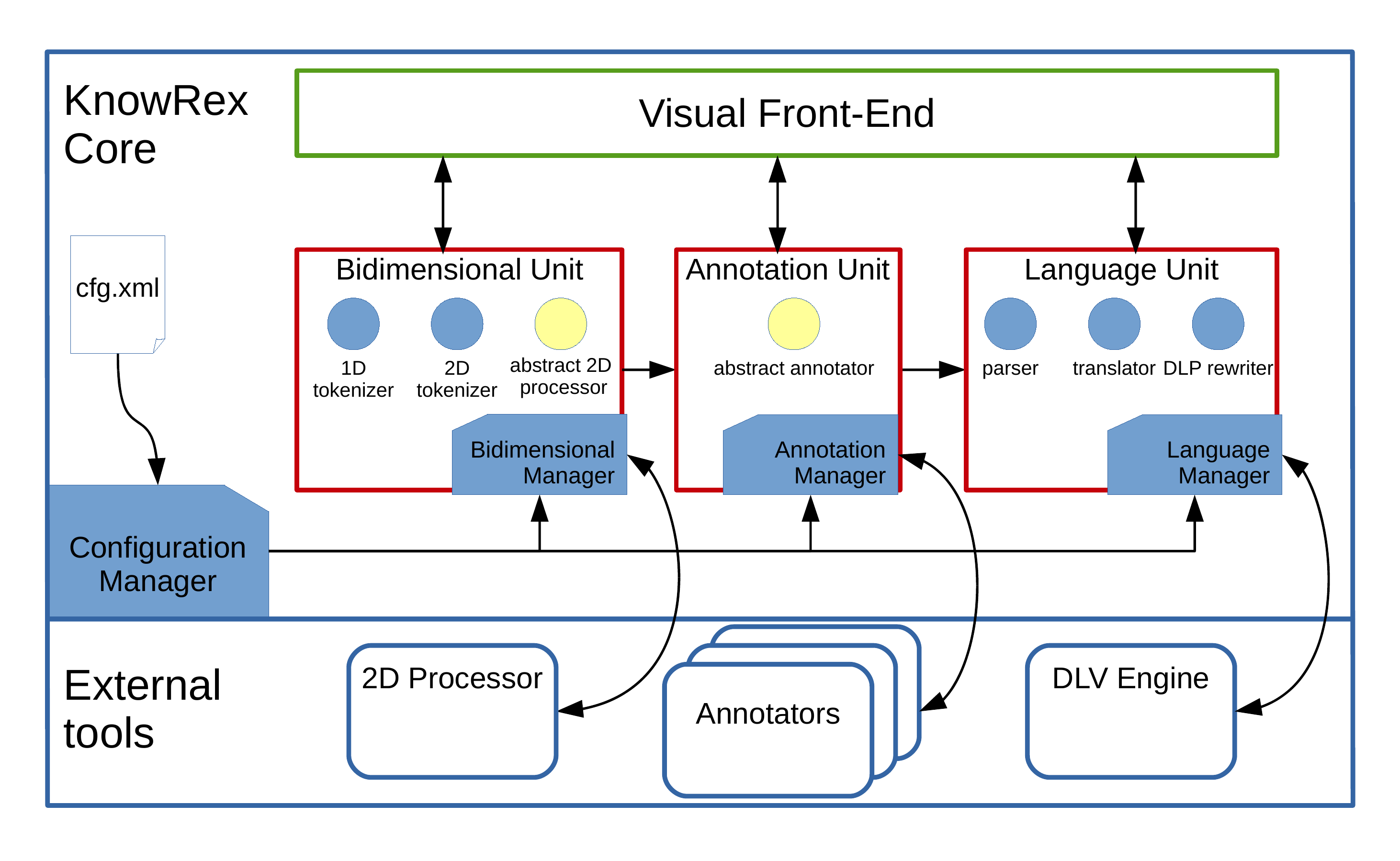}
\caption{Architecture of the KnowRex system for Semantic Information Extraction.}
\label{fig:knowrex-arch}
\end{center}
\end{figure*}

\knowrex consists of a core system and a set of external tools (see Fig.~\ref{fig:knowrex-arch}).
The Semantic Information Extraction is governed by three main components: Bi-Dimensional Unit, Annotation Unit and Language Unit, that are configured during the design phase, and during runtime are responsible for consecutive stages of document analysis, information extraction and processing.

The Bi-Dimensional Unit is responsible for a structural analysis of the input documents.
The more information about the expected layout, structure or typical features of the input documents is given, the better representation will be obtained, and better quality of information extraction during later stages can be achieved.
%
In the Annotation Unit, the system uses external tools to identify objects of certain classes within the document.
%
The Language Unit is responsible for high-level extraction of information.
The semantic descriptors used here work with the results of the annotation and bi-dimensional stages.
They take as input information about the structure of the document, the objects identified by semantic annotators, their placement within the document, proximity to each other etc., and build more complex objects for the object model.


\section{Automatic curriculum analysis}\label{sec:knowrex-usecase}

\begin{figure*}[!h]
\begin{center}
\includegraphics[width=\textwidth]{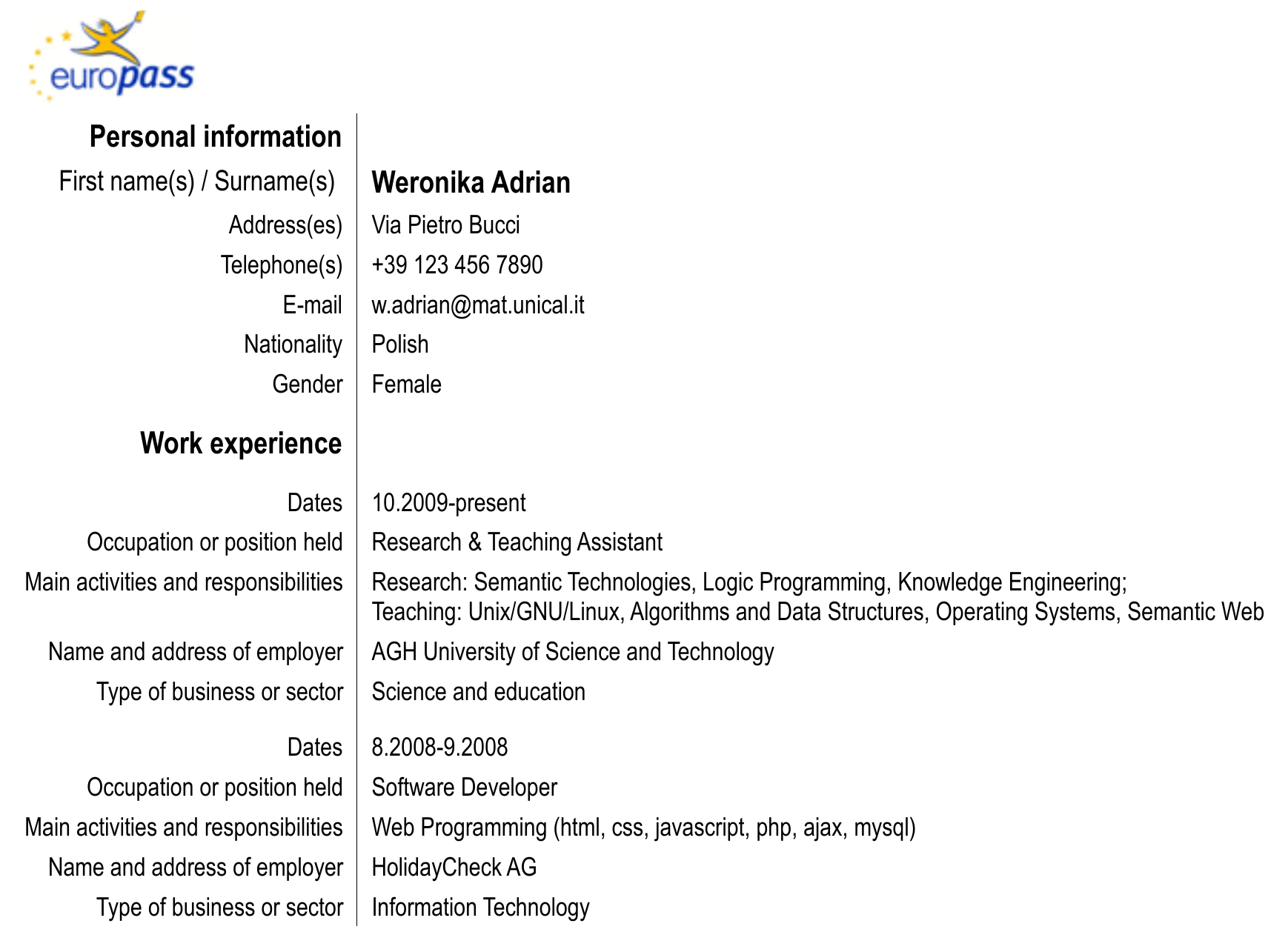}
  \caption{Fragment of a Curriculum Vitae PDF-document in the Europass style.}
  \label{fig:europass}
\end{center}
\end{figure*}

For testing our framework, we selected the European standard style
for Curriculum Vitae documents called Europass (see Figure~\ref{fig:europass}).
This choice ensures that the input documents have similar two-column
layout and organization of data.
Despite some differences between single Europass CV documents,
they can be seen as HUD,
and we can assume a certain \emph{template},
which consists in:
$(i)$ two-column layout,
$(ii)$ same file format (actually a pdf that contains no information about
sections/subsections which must be reconstructed at runtime),
$(iii)$ fixed set of labels (in the left column),
$(iv)$ common domain of values (personal information, education, work experience etc.).



%

The problem of recruiters and the goal of our system is to extract
appropriate information from a collection of documents,
and enter it into the target database.
Some information can be localized by identifying
appropriate sections and labels in the left column (e.g., name, surname, address etc.).
Also, a part from the driving license,
all the information needed for relation \textsf{candidate}
are grouped together.
For other information, it may be necessary to combine the knowledge
about the structure of the input with the semantics of data.
For instance, for work and education, it is possible to locate the institutions
by analyzing the labels in the first column and extracting
everything that follows them on the right.
However, it would be beneficial to use also the \emph{semantic annotators}
that can recognize particular phrases as names of schools
or companies to get more precise information.
Finally, some information may be dispersed through the document.
For instance, one may want to extract information about the
candidates' skills, and this may be given in several ways.
There exist dedicated sections in the Europass style,
but these are not always filled by candidates.
Thus, the skills may be also extracted from the other parts of the document.
For instance, one may recognize ``practical skills''
as gained during the work experience, and ``theoretical skills'',
if they are listed within the education section.

\section{The Design Phase in KnowRex}\label{sec:knowrex-design}

\begin{figure*}[h]
\begin{center}
\includegraphics
[scale=0.5]
{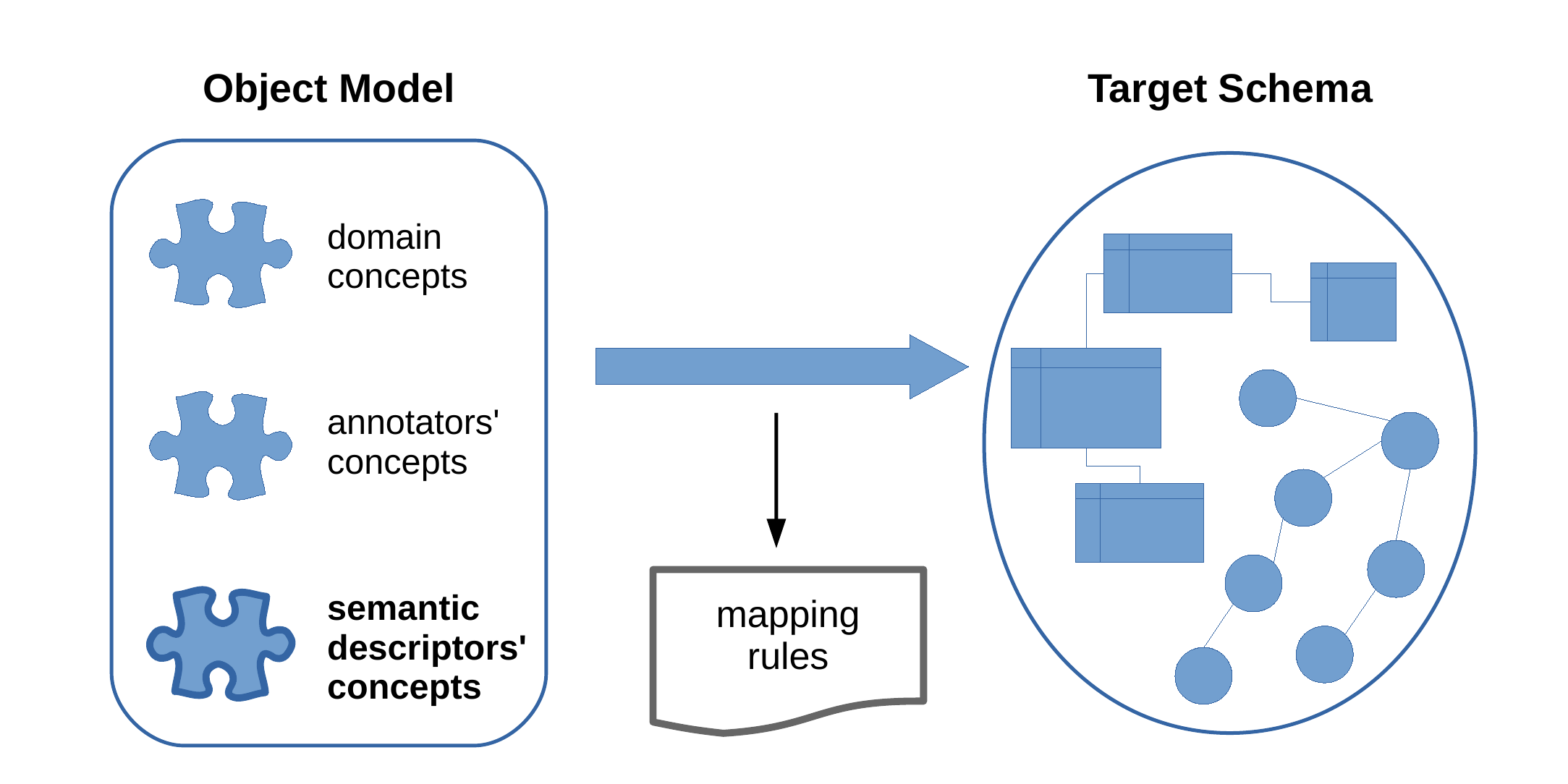}
\caption{Design Phase concepts: \emph{object model}, \emph{mapping}, and \emph{target schema}.}
\label{fig:knowrex-design}
\end{center}
\end{figure*}

During the design phase, a \knowrex project is configured to perform
operations on a collection of HUD,
to obtain information desired by user, in a particular form.
To do it, the designer should
reflect on \emph{what they have} and \emph{what the want to obtain} (see Figure~\ref{fig:knowrex-design}).
The former means identifying a \emph{template} which a vague concept for describing the common features of HUD.
A template must be formalized in the form of \emph{object model}.
Elements of it will be extracted by different components (two-dimensional processing tools, annotators, and semantic descriptors).
The \emph{target schema}, in turn, is typically formalized as an ontology or database schema.
The designer configures the system and arranges the external tools
so that the object model can be built.
Then, they write logic rules that map the object model into target schema.
The result of the design phase is used at runtime
to process the actual documents to create the semantic view.

\subsection{Definition of the Target Schema}
\label{sec:knowrex-design-semview}
%
This step is crucial for the definition of a desired output of the system.
The designer has to decide how to organize the information that
will be extracted from the documents.
The target schema
may be either for a relational database or for an ontology.
%
%
For the target database,
we may consider only
the portion of information relevant.
Let us assume the following target schema for the considered use case:
\textsf{candidate(\underline{Id}, Name, Surname, Phone, Email, Address, Gender, Nationality,
License);}
\textsf{workExperience(\underline{Id}, Company, BusinessSector, StartDate, EndDate);}
\textsf{candWE(IdCandidate, IdWorkExperience).}
The schema should be consistent and realistic i.e.,
it should be easy to populate it manually,
only by analyzing the input documents.

\subsection{Definition of the Object Model}
\label{sec:knowrex-design-om}

%
By considering the template and the target schema,
the designer fixes an object model for HUD,
which consists of a hierarchical forest-like structure.
To define it, we have used the ontology language OntoDLP~\cite{RiccaLeone07}
%
in which
one can define object types, relation types and
express relationships between objects.
Object types are preceded by keyword \textsf{\textbf{entity}},
and the subclass relationship is expressed via the term \textsf{\textbf{isa}}.
Objects may have zero or more attributes which are specified in the type definition,
by giving their names and types.
By default, a class inherits attributes from its superclass.
Relation types can be defined by keyword
\textsf{\textbf{relation}}, and giving a name and attributes
for this relation (see Section~\ref{sec:knowrex-runtime} for examples).
%

Within the object model, a few types of objects are identified.
First, there are concepts that belong to an ontology representation
of a document.
This representation is independent of the use case,
it is present in \knowrex by default and does not need configuration.
It contains one-dimensional objects such as token (basic elements of text)
and delimiters, such as start and end of a line.
It also provides two-dimensional concepts, empty and filled cells,
to represent the basic elements of the document structure.

{
\footnotesize{
\begin{lstlisting}
entity ontologyObject.
  entity oneDimObject isa ontologyObject.
    entity token isa oneDimObject. 
    entity delimiter isa oneDimObject.    
      entity startOfLine isa delimiter.
      entity endOfLine isa delimiter.
  entity biDimObject isa ontologyObject.
    entity cell isa biDimObject.
      entity emptyCell isa cell.
      entity filledCell isa cell(value:string).
\end{lstlisting}
}
}

In the second group of concepts, there are categories
that can be identified within the content of the document.
For a CV use case, we can think of places, persons, companies, schools,
skills, different professional terms, e.g. names of programming languages, languages etc.
This set of concepts is defined by a designer and heavily depends on the use case, e.g.:
\footnotesize
\begin{lstlisting}
entity semanticCategory(value:string).
  entity person isa semanticCategory.
  entity place isa semanticCategory.
  entity date isa semanticCategory.
  entity company isa semanticCategory.
  entity educationInstitution isa semanticCategory.
\end{lstlisting}
\normalsize

Finally, there is a group of concepts that describe domain-dependent elements
of the structure of the document.
These are concepts typically appearing in the considered HUD,
such as section headlines, typical labels etc.
These concepts are also specified by a designer, e.g.:
%
\footnotesize
\begin{lstlisting}
entity domainObject.
  entity eucv_label isa domainObject.
    entity eucv_name_label isa eucv_label.
    entity eucv_phone_label isa eucv_label.
    entity eucv_email_label isa eucv_label.
  entity eucv_label_box isa domainObject.
    entity eucv_name_label_box isa eucv_label_box.
   entity eucv_phone_label_box isa eucv_label_box.
\end{lstlisting}
\normalsize


\subsection{Arrangement of the Semantic Annotators}
\label{sec:knowrex-design-sa}
In this step, the designer
selects the annotators to be used, then chooses classes
that should be searched for, and configures each annotator: provides a mapping
from the tool's output to the object model, and sets the tool's specific properties.
In the case of Europass CV analysis, we have selected:
StanfordNER, 
a custom annotator for recognizing e-mail
addresses and dates, 
a dictionary-based annotator for recognizing skills defined in the 
European e-Competence Framework~\footnote{See \url{http://www.ecompetences.eu/}.},
and a label annotator based on pattern recognition
that recognizes labels typical for Europass CV.
Decisions about the arrangement of annotators are
made experimentally. 
Sometimes, it is beneficial to use more than one tool
for recognizing the same category.
The resulting potential redundancy is not harmful,
instead the recall of extraction may improve.


\subsection{Two-dimensional Document Analysis}
\label{sec:knowrex-design-2d}
Knowing the context in which certain
phrase appears is helpful for semantic information extraction.
In some input data formats, e.g. pdf documents, the information about
the structure is lost; while visible to human eye,
it is not obvious for a machine.
Thus, we need to recover the structure to obtain a meaningful
representation of the input documents.
To this end, this step configures an external
\emph{two-dimensional processor} and
a \emph{refinement module} inside \knowrexx.
As a two-dimensional processor, we have used
Quablo (\url{http://www.quablo.eu/})
that can recognize a set of regular tables within a pdf document.
The representation obtained from
this tool is then improved by a special module that works with
domain concepts, such as labels of the Europass template.
The module produces improved structure, merging appropriate cells
(for example, if a label spans across two cells, these cells will be merged).
%
Moreover, 
one- and two-dimensional tokenizers (tools inside the \knowrex core)
are used to identify the basic one- and two-dimensional objects of the document.
In the end, we obtain a grid representation of the document that consists of
two-dimensional objects (cells) containing one-dimensional ones (text fragments, delimiters).


\subsection{Semantic Descriptors Specification}
\label{sec:knowrex-design-sd}

While the semantic annotators identify single words or phrases as belonging
to specific classes (producing the ``leaves'' of the object model),
and the two-dimensional processing adds structure to the input,
the semantic descriptors can combine and use the above information
to build more complex objects.
Semantic descriptors are rules that organize two-dimensional
and one-dimensional objects into \emph{descriptions}
to extract additional information.
This is done on several levels.
To help the intuition, we illustrate the semantic descriptors by examples.

First, a designer should identify parts of the document that will help
to localize other data portions, e.g.:

\verba
\footnotesize
\begin{verbatim}
<eucv_email_label_box()> ::- <filledCell()> CONTAINS <eucv_email_label()>
\end{verbatim}
\normalsize
\verba

\noindent
With this simple descriptor, we intend to create a (two-dimensional)
concept \mytt{eucv\_email\_label\_box} that defines a cell in
which there is a (one-dimensional) \mytt{eucv\_email\_label}.
The object we want to extract always resides on the left-hand-side of the
operator ``\mytt{::-}'' (in the head of a descriptor),
while on the right (in the body), there are objects that
must be found. 
In this example, we look for a cell within which
there is a particular domain concept, an \mytt{eucv\_email\_label}.
If we find a cell with this label inside, the cell can be
recognized as a \mytt{eucv\_email\_label\_box}.

Descriptors can join several cells that appear in a document one after
another (horizontally or vertically).
This is useful, if we want to say that there exist a particular object,
if there is a specific sequence of cells, e.g.:

\verba
\footnotesize
\begin{verbatim}
<candidateEmail(E)> ::- <eucv_email_label_box()>
                        (<filledCell(X)> CONTAINS <email(X)> {E:=X;})
\end{verbatim}
\normalsize
\verba

\noindent
This description should be read as: ``A \mytt{candidateEmail} is
a two-dimensional object that captures two cells: the first is
an \mytt{eucv\_email\_label\_box} and is followed (horizontally)
by a \mytt{filledCell} that contains a (one-dimensional)
object \mytt{email} with value \mytt{X}.
The new object spans across both cells, and the value of the object
\mytt{email} becomes the value of \mytt{candidateEmail}.''
By using the context (first there is a box with an e-mail label,
and then there is a cell with an e-mail address),
we ensure that, even if the CV contains a few e-mail addresses,
we select the correct one, because if the e-mail address appears in this place,
it must be in the Personal Information section, and thus
it is the e-mail of the candidate.

We can also aggregate the concepts and attributes extracted by other semantic
descriptors to build more complex ones, e.g.:

\verba
\footnotesize
\begin{verbatim}
<personalInformation(N, S, A, P, E, Nt G)> ::|
                <candidateName(X)> {N:=X;} <candidateSurname(X)> {S:=X;}
               <candidateAddress(X)> {A:=X;} <candidatePhone(X)> {P:=X;}
         <candidateNationality(X)> {Nt:=X;} <candidateGender(X)> {G:=X;}
\end{verbatim}
\normalsize
\verba

\noindent
This semantic descriptor aggregates results of other descriptors
that extract single information about a candidate.
It describes a sequence of concepts that must appear one after another vertically,
which we mark with the ``\mytt{::|}'' operator.
%

Within cells, we can create one-dimensional descriptors by using the operator ``\mytt{::}''.
A recurrence structure ``\mytt{(\textit{sequence of terms})+}''
and a keyword ``\mytt{...}'' that allows to skip irrelevant data allow to build complex descriptions such as:

\verba
\footnotesize
\begin{verbatim}
<list_of_skills(S)> :: {S:=[];} <startOfLine> ...
                       (<IndustryTerm(S1)> {S&=S1;} ...)+ <endOfLine>
\end{verbatim}
\normalsize
\verba

\noindent
This descriptor works on one-dimensional objects that are all located in one cell
(treated as a single line thanks to the two-dimensional processing).
Here, we want to create a list, so we initialize the attribute \mytt{S:=[]}.
Then, we look for a concept \mytt{IndustryTerm},
recognized by a semantic annotator, append
its attribute value to \mytt{S} (\mytt{\{S\&=S1;\}}),
and place the term in a recurrence structure.
The expression \mytt{(<IndustryTerm(S1)>\{S\&=S1;\} ...)+} means that there may
be some objects after the \mytt{IndustryTerm} that we ignore, and if we find another
object \mytt{IndustryTerm}, we append its attribute value to the list again.
By using the keyword ``\mytt{...}'' before the recurrence, we say that
we can skip some objects, i.e., the recurrence structure may appear anywhere
between the \mytt{startOfLine} and the \mytt{endOfLine}.
The descriptor creates a new object \mytt{list\_of\_skills}
that stores as an attribute a list of \mytt{IndustryTerm} objects' attributes.

Finally, semantic descriptors may use the information about the placement of
objects within the document (e.g. presence of a given object within specific section)
to extract new objects that are not explicitly defined in text, e.g.:

\verba
\footnotesize
\begin{verbatim}
<list_of_practical_skills(S)> ::- <eucv_work_act_resp_label_box()>
                   (<filledCell(X)> CONTAINS <list_of_skills(X)> {S:=X;})
\end{verbatim}
\normalsize
\verba

\noindent
In this example, we use a \mytt{eucv\_work\_act\_resp\_label\_box},
a domain concept
that represents a cell containing
``Activities and Responsibilities'' label for selected Work Experience.
This way, we look for lists of skills present only within the Work Experience
subsections (and not for example within Education ones) and we can
call them practical skills.

\subsection{
Defining a Mapping from Object Model to Target Schema} 
\label{sec:knowrex-design-mapping}
The design phase in \knowrex is completed with the definition of a mapping from object
model classes to the concepts of the target schema.
This mapping, written in a form of Datalog rules, is used to automatically create
a semantic view of the (structured) input documents during the runtime phase.
In the head of rules, there are concepts from the target schema, and in
the body -- objects from the object model (and auxiliary objects such as candidate ID).
Partial mapping of the object model 
to the target schema 
is as follows:

\verba
\footnotesize
\begin{verbatim}
candidate(Id,N,S,P,E,A,G,Nt,D,L) :- ID:cv_candidate_id(Id),
           PI:personalInformation(N,S,A,P,E,Nt,D,G),
           CDL:candidateDrivingLicence(L).
workExperience(WExpId, Company, BusinessSector, Start, End) :-
           C:company(WExpId,Company,BusinessSector),
           WED:workExperienceDates(WExpId,Start,End).
\end{verbatim}
\normalsize
\verba

\noindent 
%
When the design phase ends, all the decisions are saved in the configuration files of the system.
The semantic descriptors are translated into logic rules. 

\section{The Runtime Phase of the System}\label{sec:knowrex-runtime}
\begin{figure*}[h]
\begin{center}
\includegraphics
[width=\textwidth]%
{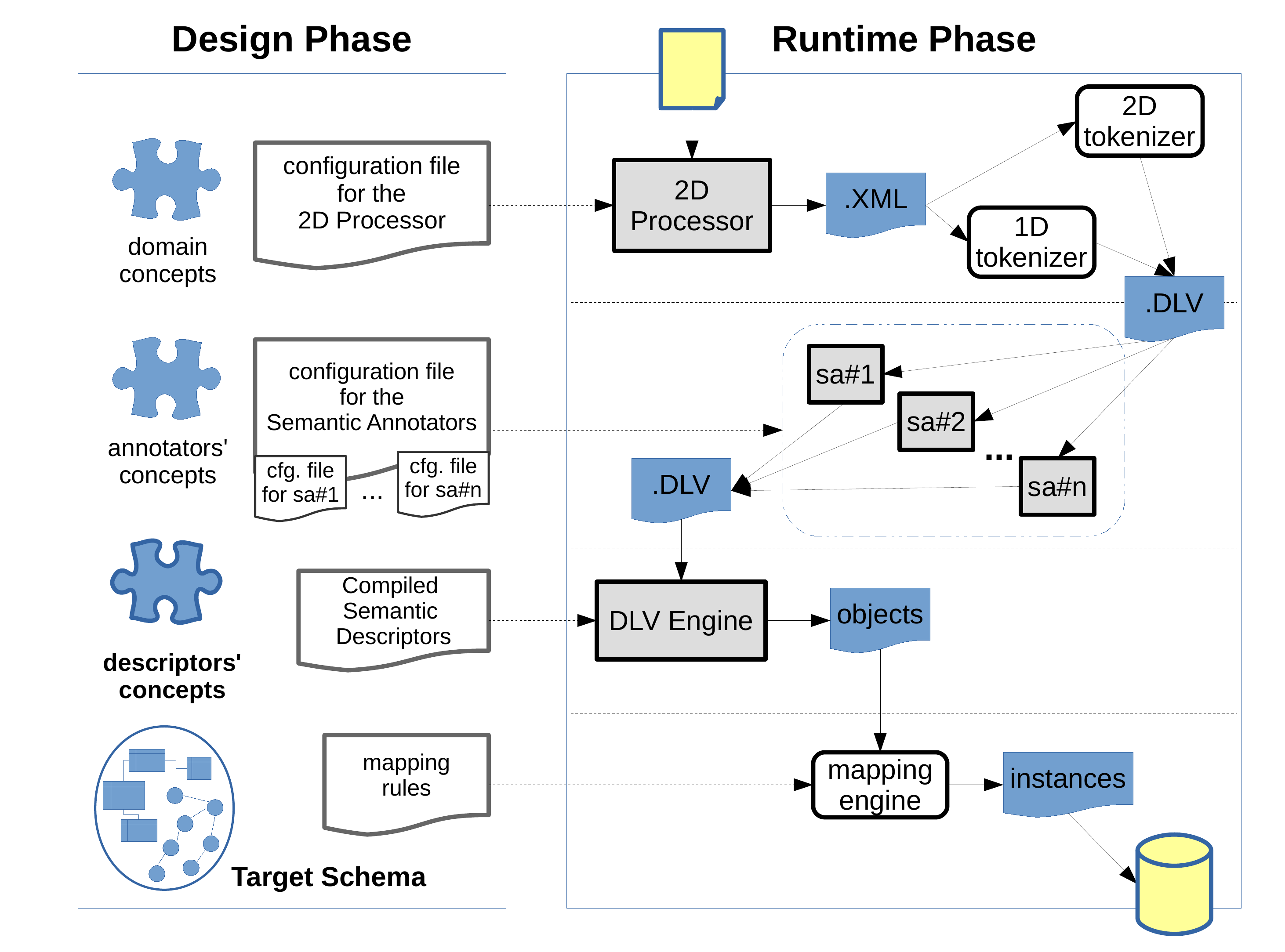}
\caption{Runtime Phase of \knowrex system.}
\label{fig:knowrex-runtime}
\end{center}
\end{figure*}

Once the design of the project is done, KnowRex can be run over a collection of input documents.
The flow of the operations and the relations between the design and runtime phases may be observed in Figure~\ref{fig:knowrex-runtime}.

In the stage of the document analysis,
%
first, a two-dimensional processor is used.
Its output is then refined according to
domain knowledge (specific labels, structure elements or keywords).
%
This improved structure is analyzed by one- and two-dimensional
tokenizers, tools hidden from a user, that identify the atomic one- and
two-dimensional components of a document (tokens and cells).
A logical fact base is obtained that represent the document as a two-dimensional ``grid''.
%
\knowrex uses a two-dimensional representation of objects
that helps localize them within the documents.
The definitions of the position relations in OntoDLP are as follows:
\footnotesize
\begin{lstlisting}       
relation position(obj:ontologyObject, start:int, end:int).
  relation onePosition(obj:oneDimObject, start:int, end:int).
  relation biPosition(obj:biDimObject, xstart:int, ystart:int, xend:int, yend:int).
relation belongsTo(obj:oneDimObject, obj2:biDimObject)      
\end{lstlisting}
\normalsize

At the end of the two-dimensional processing stage,
an ontological model of the document is obtained.
It contains information about positions of the one- and two-dimensional
objects within the document.
For each two-dimensional object,
a relation \textsf{biPosition} is added that specify the row
and column on the document ``grid'', where the object appears, e.g.:
\footnotesize
\begin{verbatim}
filled19:filledCell('anna@w3.org'). biPosition(filled19,1,8,2,9).
\end{verbatim}
\normalsize
For all one-dimensional objects (that are located inside the two-dimensional cells),
two relations are added: \textsf{belongsTo}
that identifies the containing cell by its id,
and \textsf{onePosition} which denotes the position of the object within a cell:
\footnotesize
\begin{verbatim}
tk123:token('manager').onePosition(tk123,0,6).belongsTo(tk123,filled80).
tk124:token('of').onePosition(tk124,6,7).belongsTo(tk124,filled80).
\end{verbatim}
\normalsize

\noindent This representation is \emph{normalized} i.e.,
the positions of blank spaces are omitted and the tokens
follow one another.
Such a representation is a reference for
semantic annotators that may treat blank spaces differently.


Then comes the annotation stage, in which selected semantic annotators
are run over the identified cells and label the parts of text
as objects belonging to different classes (such as
Places, Persons, IndustryTerms, etc.)
The representation of the identified objects (new logic
facts that carry information about the annotator that found
the object) is added to the fact base, e.g.:
\footnotesize
\begin{verbatim}
annS2:email('anna@w3.org').one_position(0,10).belongs_to(annS2,filled19).
\end{verbatim}
\normalsize

Once the annotation stage is finished, the semantic descriptors
which have been compiled into logic rules are executed over the facts
representing the objects within a document.
Each descriptor is transformed into a set of logical rules
that first extract the portion of the document complying to the descriptor body,
and then create a new object, specified in the descriptor head.

Each descriptor is internally represented as an automaton.
After setting the initial configuration, 
each element of a descriptor is treated as a transition that allows to go from one state to the next one.
The condition that one object must appear after another in a document is realized by checking the positions of the objects using \texttt{biPosition} and \texttt{onePosition} relations.
The relation \texttt{belongsTo} checks the conditions expressed by the \texttt{CONTAINS} keyword.
The attributes of the objects are passed between the rules by using variables.

For instance, the descriptor from Example 2 in Section~\ref{sec:knowrex-design-sd}:
 {\footnotesize
 \begin{verbatim}
 <candidateEmail(E)> ::- {E:='';} <eucv_email_label_box()>
                <filledCell(X)> CONTAINS <email(X)> {E:=X;} 
\end{verbatim} }
\noindent is translated into the following logic rules: 
 \begin{enumerate}
 \item Extracting candidate email from the document:
 {
 \footnotesize{
 \begin{verbatim}
 init_conf_candidateEmail(0,"").
 conf_candidateEmail(1, "", Xe, Ys, Xe_1, Ye) :- 
     init_conf_candidateEmail(0,""), 
     eu_cv_email_label_box(Id, Lc4), 
     bi_position(Id, Xe, Ys, Xe_1, Ye).
 conf_candidateEmail(2, Lc1, Xs, Ys, Xe_1, Ye) :- 
     conf_candidateEmail(1, Gl3, Xs, Ys, Xe, Ye), 
     filledCell(Id, Lc1), bi_position(Id, Xe, Ys, Xe_1, Ye).
 conf_candidateEmail(3, Lc2, Xe, Ys, Xe_1, Ye) :- 
     conf_candidateEmail(1, Gl3, Xs, Ys, Xe, Ye), 
     filledCell(Id, Lc1), bi_position(Id, Xe, Ys, Xe_1, Ye), 
     email(IdContains, Lc2), belongs_to(IdContains,Id).
 \end{verbatim}
 }
 }
 \item Creating a new object for the object model, with its position:
 {
 \footnotesize{
 \begin{verbatim}
 aux_candidateEmail(AutoGen,Gl3, Xs, Ys, Xe, Ye) :- 
     conf_candidateEmail(3,Gl3, Xs, Ys, Xe, Ye), AutoGen=#newID.

 AutoGen : candidateEmail(Gl3) :- 
       aux_candidateEmail(AutoGen,Gl3, Xs, Ys, Xe, Ye).
 bi_position(AutoGen, Xs, Ys, Xe, Ye) :-
       aux_candidateEmail(AutoGen,Gl3, Xs, Ys, Xe, Ye).
 \end{verbatim}
 }
 }
These 
rules use the output of the extraction and create an object in OntoDLP, together with its one- or bi-dimensional position (and optionally, \emph{belonging to} a cell, if it is a one-dimensional object).
\end{enumerate}

\begin{figure*}[h]
\begin{center}
\includegraphics
[width=\textwidth]%
{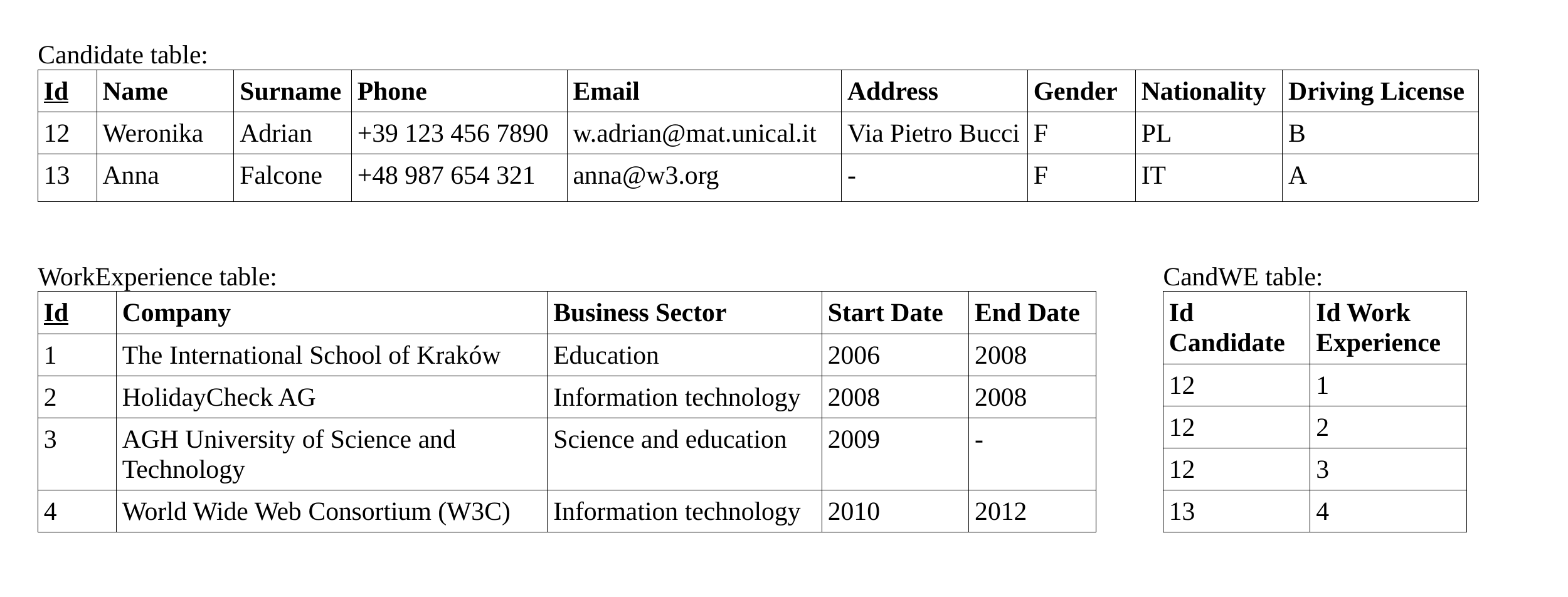}
\caption{Table output (fragment of the semantic view) of the input documents.}
\label{fig:knowrex-output}
\end{center}
\end{figure*}

Finally, the extracted objects are transformed
into the instances of the semantic view (see Figure~\ref{fig:knowrex-output})
with use of the mapping defined in the design phase. 
%
Technically, this is done by additional logic rules that create instances for the target representation from the objects (in OntoDLP) of the object model.

\section{Discussion and Conclusion}\label{sec:concl}
We have described an ontology-based approach
for extracting and organizing
semantically rich information from HUD.
This approach has been implemented in a system called KnowRex, which
has been tested on curricula in the Europass style, stored as pdf files.
Roughly, the design phase has been carried out in two man-weeks.
From our preliminary analysis over 80 CVs,
it appeared that the two-dimensional structure recognition
and the recall of third-party annotators are the main bottlenecks.
With initial configuration for the Europass template,
Quablo worked well for about 50\% of documents.
For further 20\% of documents, satisfying results were obtained
by small adjustments of the tool (margin toleration etc.).
While the precision of the semantic annotators is satisfying, their
sometimes low recall may be compensated by adjusting dictionary-based annotators.
Logical rules (semantic descriptors and mapping rules)
worked as expected on the found objects without loss of precision.
KnowRex is sufficiently flexible and modular to be suitable for
various scenarios in which HUD is available.
%

\section*{Acknowledgement}
The work has been supported by the Calabrian Region within the project "THT~-- Talent Hunter Technology" (CUP: J84E07000540005).

\label{sec:biblio}
\bibliographystyle{splncs03}
\bibliography{knowrex_biblio}

\end{document}